\documentclass[dvipsnames,format=sigconf,nonacm,anonymous=false]{acmart}

\AtBeginDocument{
  }

\usepackage{subcaption}
\usepackage{multirow}
\usepackage{array}
\usepackage{setspace}

\begin{document}

\title{Geometric Learning in Black-Box Optimization: A GNN Framework for Algorithm Performance Prediction}

\author{Ana Kostovska}
\orcid{0000-0002-5983-7169}
\affiliation{
  \institution{Jo\v{z}ef Stefan Institute}
  \city{Ljubljana} 
 \country{Slovenia}
}

\author{Carola Doerr} 
\orcid{0000-0002-4981-3227}
\affiliation{
  \institution{Sorbonne Universit\'e, CNRS, LIP6}
  \streetaddress{}
  \city{Paris} 
  \country{France}}

\author{Sa\v{s}o D\v{z}eroski}
\orcid{0000-0003-2363-712X}
\affiliation{
  \institution{Jo\v{z}ef Stefan Institute}
  \city{Ljubljana} 
 \country{Slovenia}
}

\author{Pan\v{c}e Panov}
\orcid{0000-0002-7685-9140}
\affiliation{
  \institution{Jo\v{z}ef Stefan Institute}
  \city{Ljubljana} 
 \country{Slovenia}
}

\author{Tome Eftimov}
\orcid{0000-0001-7330-1902}
\affiliation{
  \institution{Jo\v{z}ef Stefan Institute}
  \streetaddress{}
  \city{Ljubljana} 
 \country{Slovenia}
}

\renewcommand{\shortauthors}{Kostovska et al.}

\begin{abstract}

Automated algorithm performance prediction in numerical black-box optimization often relies on problem characterizations, such as exploratory landscape analysis features. These features are typically used as inputs to machine learning models and are represented in a tabular format. However, such approaches often overlook algorithm configurations, a key factor influencing performance. The relationships between algorithm operators, parameters, problem characteristics, and performance outcomes form a complex structure best represented as a graph.

\begin{sloppypar}

    This work explores the use of heterogeneous graph data structures and graph neural networks to predict the performance of optimization algorithms by capturing the complex dependencies between problems, algorithm configurations, and performance outcomes. We focus on two modular frameworks, modCMA-ES and modDE, which decompose two widely used derivative-free optimization algorithms: the covariance matrix adaptation evolution strategy (CMA-ES) and differential evolution (DE). We evaluate 324 modCMA-ES and 576 modDE variants on 24 BBOB problems across six runtime budgets and two problem dimensions.   
\end{sloppypar}

Achieving up to 36.6\% improvement in $MSE$ over traditional tabular-based methods, this work highlights the potential of geometric learning in black-box optimization.

\end{abstract}



\maketitle

\section{Introduction}

Numerical black-box optimization has long been an active area of research, resulting in the development of numerous optimization algorithms. Consequently, meta-learning tasks such as automated algorithm performance prediction~\cite{trajanov2021explainable}, algorithm configuration~\cite{schede2022survey}, and algorithm selection~\cite{kerschke2019automated} have become increasingly important for identifying the most suitable algorithm for a given problem.

Over the years, many novel algorithmic ideas have been introduced, along with incremental improvements to well-known derivative-free optimization algorithms such as the Covariance Matrix Adaptation Evolution Strategy (CMA-ES)~\cite{hansen1996_cmaes} and Differential Evolution (DE)~\cite{storn1997_de}. These advancements have further diversified the landscape of optimization algorithms, making it more challenging to systematically compare and select among them.

\begin{sloppypar}
    To address this, modular optimization frameworks such as modCMA-ES~\cite{nobel_modcma_assessing} and modDE~\cite{DBLP:journals/corr/abs-2304-09524} decompose CMA-ES and DE into modular components, enabling researchers to analyze how different configurations and parameter settings affect performance. By capturing detailed information about algorithmic structure and interactions, these frameworks offer a rich resource for algorithm analysis. However, despite their potential, they remain underutilized in meta-learning tasks.
\end{sloppypar}

Traditional approaches to algorithm performance prediction typically rely on tabular representations of problem features, such as Exploratory Landscape Analysis (ELA)~\cite{mersmann_exploratory_2011}. While effective, these methods fail to capture the complex inter-dependencies among problems, algorithms, configurations, and performance outcomes. Representing these entities as nodes in a graph, with edges capturing their relationships, naturally reflects the relational structure of the data and supports richer modeling.

Geometric learning, a branch of machine learning that handles data in non-Euclidean domains like graphs and manifolds~\cite{bronstein2017geometric}, provides a suitable framework for this task. Its approaches range from shallow node embeddings to more advanced Graph Neural Networks (GNNs), all leveraging relational and topological information within the data.

An early demonstration of this concept involved modeling performance prediction as a link prediction task. Specifically, in our previous work~\cite{kostovska2023using}, we employed shallow node embeddings to infer ``missing'' performance links between modular optimization algorithms and black-box problems, labeling a link as solved if an algorithm reached a predefined precision threshold within a fixed budget of function evaluations. While effective, this approach framed performance prediction as a binary classification task rather than a more expressive regression problem. Moreover, shallow embeddings restricted the framework to transductive learning, limiting generalization to unseen problems.

GNNs address these limitations by capturing higher-order dependencies and enabling inductive learning. While not yet applied to modular optimization algorithms, GNNs have been successfully used in related domains. For example, Lukasik et al.~\cite{lukasik2021neural} employed GNNs to predict neural architecture performance on NAS-Bench-101~\cite{ying2019bench}, and Singh et al.~\cite{singh2021using} modeled deep networks as graphs to improve runtime prediction. Chai et al.~\cite{chai2023perfsage} proposed PerfSAGE, a GNN-based model that predicts inference latency, energy usage, and memory footprint for neural networks on edge devices.

\noindent\textbf{Our contributions:} In this study, we apply GNNs for predicting the performance of modular optimization algorithms. First, we propose a heterogeneous graph representation that encodes data from \textit{modCMA-ES} and \textit{modDE}, capturing problems, algorithm components, parameters, and performance in a unified graph structure. Second, building on this representation, we introduce a message-passing GNN framework designed to operate on the heterogeneous graph, performing node regression on performance nodes to predict an algorithm’s performance on a given problem. Finally, we present experiments demonstrating that incorporating relational structure via graph-based models improves predictive performance compared to conventional tabular approaches.

\noindent\textbf{Outline:} The remainder of the paper is structured as follows: Section~\ref{sec:methodolgy} details the methodology employed in our study, including the graph construction and GNN design. Section~\ref{sec:experimental_design} outlines the experimental setup. Section~\ref{sec:results} presents the study's results, and Section~\ref{sec:conclusions} concludes with insights and directions for future work.

\begin{sloppypar}
    \noindent\textbf{Code and Data:} All related code and data used in this study are available at: \url{https://github.com/KostovskaAna/GNN4PerformancePrediction}.
\end{sloppypar}

\section{Methodology}
\label{sec:methodolgy}
This section outlines our approach to predicting algorithm performance using GNNs on benchmarking data from modular optimization frameworks. We describe the heterogeneous graph representation, followed by the GNN model architecture.

\begin{figure}
    \centering
    \includegraphics[width=0.85\linewidth]{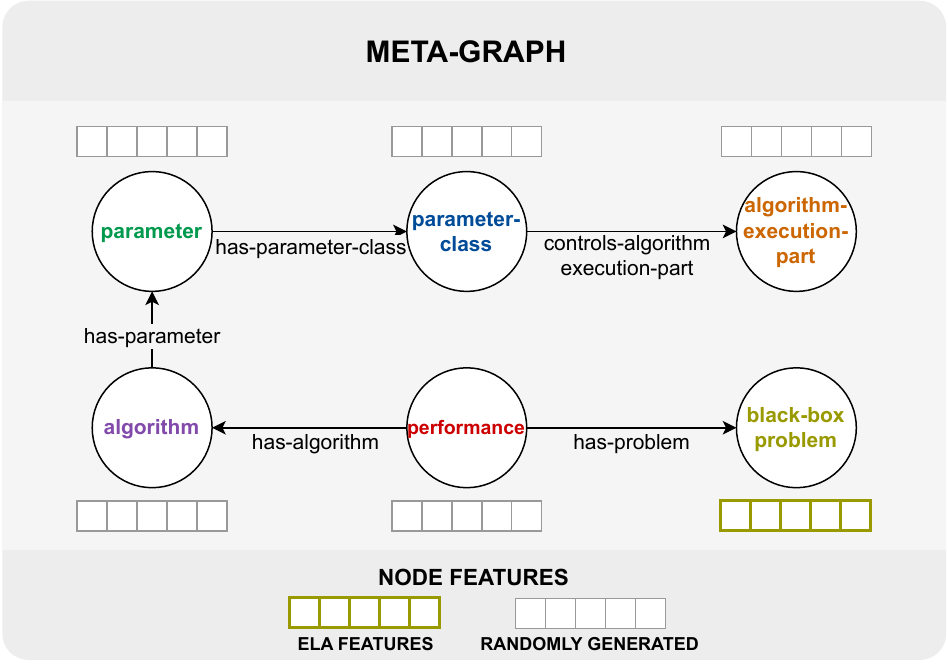}
    \caption{The meta-graph for the heterogeneous graph. }
    \label{fig:metagraph}
\end{figure}

\begin{figure}
    \centering
    \includegraphics[width=0.85\linewidth]{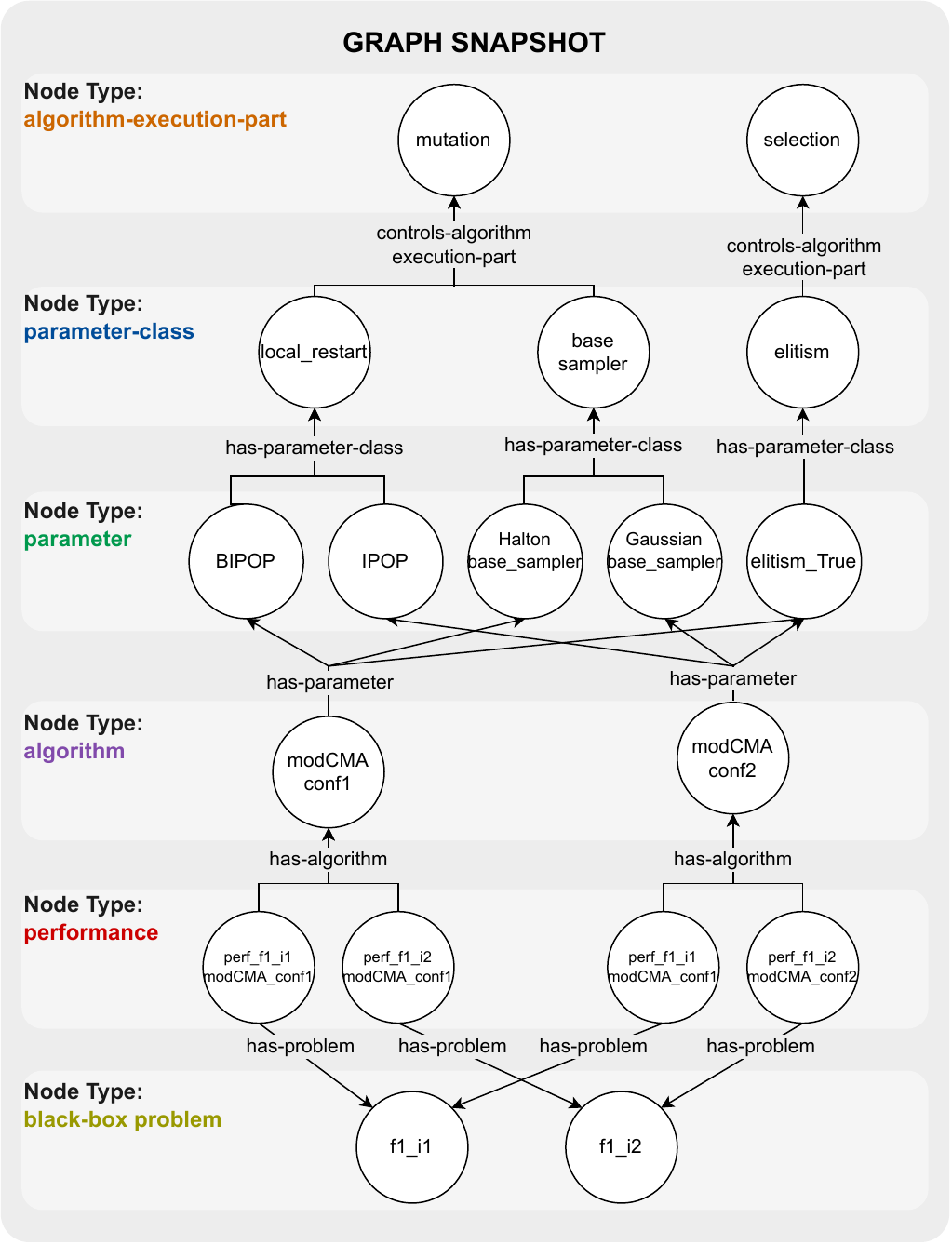}
    \caption{An illustration of an instantiation of the meta-graph, showing a snapshot of the heterogeneous graph.}
    \label{fig:graph-example}
\end{figure}

\subsection{Graph representation}
We use a heterogeneous graph to capture the black-box optimization problems, modular optimization algorithms, their configuration, and corresponding performance scores. 

\begin{sloppypar}
    A \textit{heterogeneous graph} (HG) is defined as a graph $HG= \{ \mathcal{V}, \mathcal{E}, \mathcal{R}, \mathcal{T} \}$, where $\mathcal{V}$ represents the set of nodes, $\mathcal{E}$ represents the set of edges, $\mathcal{R}$ represents the set of relation types, and $\mathcal{T}$ represents the set of node types. Each node $v \in \mathcal{V}$ is associated with a node type through a mapping function $T(v): \mathcal{V} \to \mathcal{T}$, and each edge $e \in \mathcal{E}$ is associated with a relation type through a mapping function $R(e): \mathcal{E} \to \mathcal{R}$. For the graph to be considered heterogeneous, the sum of distinct node and relation types must be greater than two, i.e., $|\mathcal{T}| + |\mathcal{R}| > 2$. In addition to type mappings, each node $v \in \mathcal{V}$ is associated with a feature vector $\mathbf{x}_v \in \mathbb{R}^d$, where $d$ denotes the dimensionality of the feature space. These node features encode the properties of the nodes.
\end{sloppypar}

In Figure~\ref{fig:metagraph}, we illustrate the meta-graph of the heterogeneous graph that we propose. This \emph{meta-graph} serves as a higher-level template for our \(HG\), representing \(\mathcal{T}\) (the set of node types) as nodes and \(\mathcal{R}\) (the set of relation types) as edges between them. It includes six node types: \emph{parameter}, \emph{parameter class}, \emph{algorithm execution part}, \emph{algorithm}, \emph{performance}, and \emph{black-box optimization problem} and five edge (relation) types: \emph{has-parameter}, \emph{has-parameter-class}, \emph{controls-algorithm-execution-part}, \emph{has-algorithm}, and \emph{has-problem}.

In Figure~\ref{fig:graph-example}, we illustrate a concrete instantiation of the meta-graph for a subset of \(\mathcal{V}\) and \(\mathcal{E}\), providing a snapshot of the corresponding heterogeneous graph. It shows two black-box problems, two algorithm variants, and their associated performance nodes and parameters. The parameters belong to \emph{local\_restart}, \emph{base\_sampler}, and \emph{elitism} classes, affecting different parts of algorithm execution. We construct one graph per unique combination of problem dimensionality, runtime budget, and algorithm type, resulting in multiple graph instances.

\subsection{GNN architecture design}
We employ a heterogeneous message-passing GNN architecture to handle multiple node and relation types in our graph. In essence, message passing in a GNN involves each node gathering ``messages'' from its neighbors, transforming and combining them to update its own representation. In a heterogeneous setting, different node and edge types require distinct message-passing functions or aggregation procedures, a well-established approach in the literature~\cite{zhang2019heterogeneous}. Consequently, each relation type \(r\) dictates how messages flow from source nodes of type \(t_u\) to destination nodes of type \(t_v\). After aggregating messages for each relation, we consolidate the outputs across all relevant relation types to form the updated node embedding. Repeating this process over multiple layers enables nodes to capture information from increasingly distant parts of the graph.

\begin{figure}
    \centering
    \includegraphics[width=\linewidth]{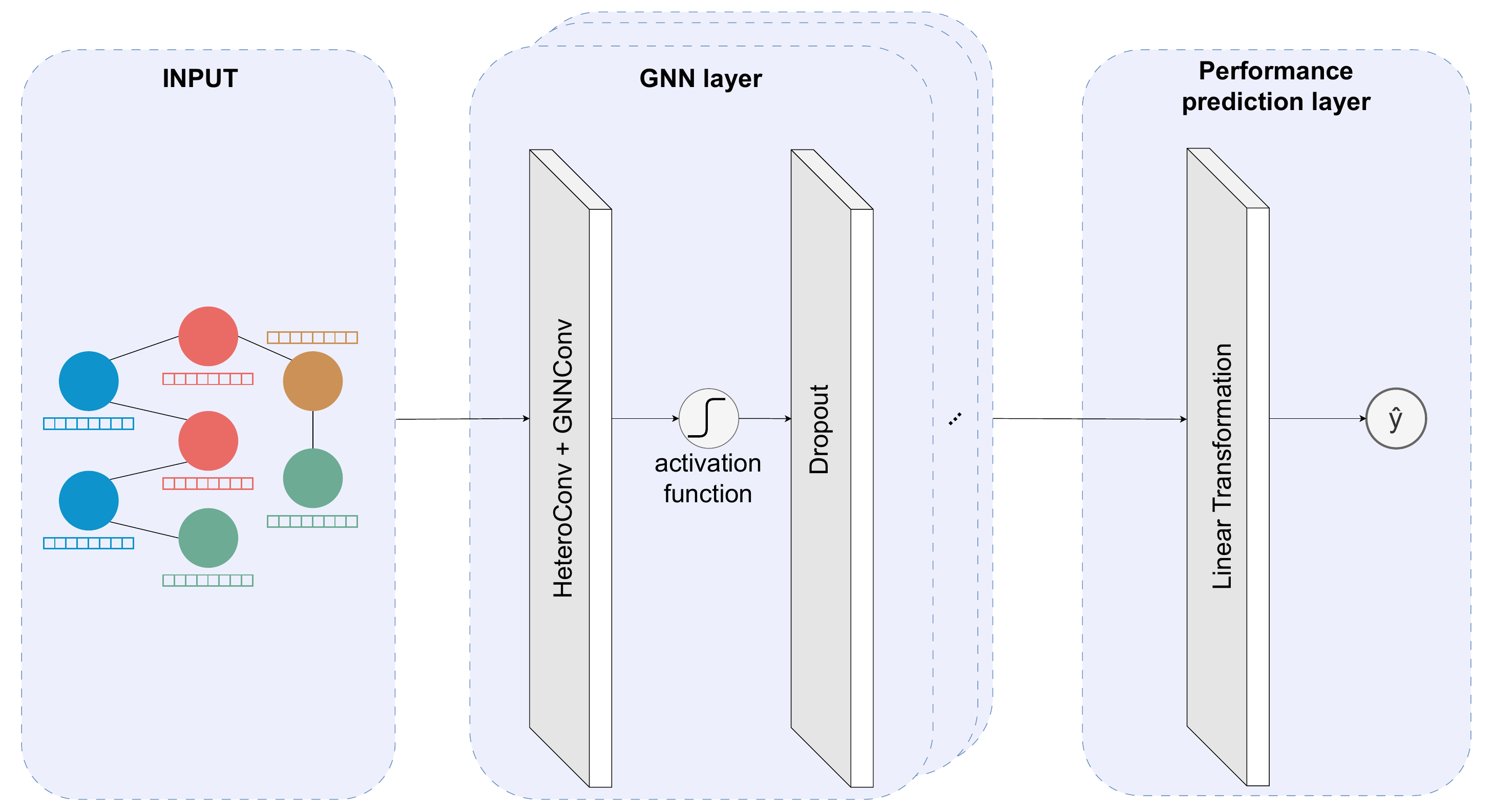}
    \caption{ An overview of the message-passing GNN architecture for predicting algorithm performance using heterogeneous graphs. }
    \label{fig:GNN_architecture}
\end{figure}

Our architecture supports stacking multiple message-passing layers (see Figure~\ref{fig:GNN_architecture}). We use GraphSAGE~\cite{hamilton2017inductive} for its inductive capabilities, though other message-passing layers are also compatible. Each layer performs relation-specific aggregation followed by cross-relation aggregation, with non-linearity introduced via an activation function and dropout applied to prevent overfitting.

The learning task is node regression, where the goal is to predict the performance of an algorithm on a specific problem instance. Each performance node is associated with a numerical score, and the GNN learns to predict it by leveraging the structure and features of the heterogeneous graph. The final GNN embeddings are passed through a linear layer to predict the performance value \( \hat{y} \).

Note that, such trained model is capable of generalizing across different algorithm variants, predicting the performance of \textit{all modular algorithm variants} for a given runtime budget and problem dimensionality. Also, although the input graph (Figure~\ref{fig:graph-example}) is directed, we add reverse edges, effectively treating the graph as undirected. This enables bidirectional information flow and improves representation learning. 

\section{Experimental setup}
\label{sec:experimental_design}

\subsection{Data collection}
\subsubsection{Problem instance portfolio and landscape data.}
We use the 24 single-objective, noiseless, continuous black-box problems from the BBOB benchmark suite from the COCO environment~\cite{hansen2020coco}. Each problem includes multiple instances generated through transformations such as offsets and scalings. We select the first five instances pre-generated by COCO for each problem at dimensions \(D=5\) and \(D=30\). To represent the problem landscapes, we use 46 ELA features available through the \emph{OPTION KB}~\cite{option_kostovska}.

\subsubsection{Algorithm instances and configuration.}
We consider two modular algorithms: \emph{CMA-ES} and \emph{DE}. For CMA-ES, we use the modCMA-ES framework~\cite{nobel_modcma_assessing}, which includes variations in sampling distributions, weighting schemes, and restart strategies. For DE, we employ the modDE package~\cite{DBLP:journals/corr/abs-2304-09524}, which supports a range of mutation operators, base selection options, crossover strategies, and update mechanisms inspired by state-of-the-art DE variants. Full details on the modules and parameter spaces are provided in~\cite{kostovska2023using}, from which this data has been reused.

In total, we analyze 324 modCMA-ES and 576 modDE variants. Each is evaluated on six function evaluation budgets, \(B \in \{50D, 100D, 300D, 500D, 1000D, 1500D\}\), where \(D \in \{5, 30\}\). The best precision (i.e., distance to the optimum) achieved at each budget is recorded.

All data used in this study is reused from Kostovska et al.\cite{kostovska2023using} and accessed through the \emph{OPTION KB}\cite{option_kostovska}.

\subsection{Model training and evaluation}

We implement the GNN architecture using the Deep Graph Library (DGL)~\cite{wang2019deep} to predict the performance of modular optimization algorithms. The model uses \texttt{SageConv} layers, based on the GraphSAGE algorithm~\cite{hamilton2017inductive}, for per-relation message aggregation. These embeddings are then combined using \texttt{HeteroConv} layers for cross-relation aggregation. Mean aggregation is applied within relations, while summation is used across relations.

We use a 4-layer GNN to capture all relevant algorithmic information. Problem nodes are represented by 46 ELA features, while features of other node types are initialized using Kaiming uniform distribution. We use GELU~\cite{hendrycks2016gaussian} to introduces non-linearity.

We perform nested cross-validation to tune dropout rates (0.1, 0.2, 0.3) and embedding sizes (32, 64, 128), repeating each experiment 10 times. Models are trained for 200 epochs using L1 loss, optimized with Adam~\cite{kingma2014adam} (learning rate 0.1, halved every 20 stagnant epochs). We adopt the same \emph{leave-instance-out nested cross-validation} protocol as in our previous work~\cite{kostovska2024using}, where we train Random Forest (RF) regressors on the 46 ELA features. These serve as baselines for evaluating our GNN models.

\section{Results }
\label{sec:results}
Building on the setup above, we conducted experiments to predict the performance of modular algorithms. Table~\ref{tab:GNN_MSE_scores_regression_models} presents MSE scores across two dimensionalities and six budgets, comparing GraphSAGE models with RF baselines using only tabular ELA features from Kostovska et al.~\cite{kostovska2024using}.

GNNs generally outperform RF across all settings. For modCMA and \(5D\) problem dimensionality, GNN improves MSE by \(0.03\)--\(0.06\) for lower budgets (\(50D\), \(100D\)), with the margin growing for larger budgets (e.g., \(5.19 \rightarrow 4.57\) at \(1500D\)). At \(30D\) problem dimensionality, gains are even more pronounced for mid to high budgets—for instance, at \(100D\), MSE drops from \(0.27\) to \(0.19\), and at \(1000D\), the relative improvement reaches \(\mathbf{36.6\%}\).

For modDE at \(5D\), GNN shows lower MSE than RF for most budgets, with the largest improvement at \(300D\) and \(500D\). At \(30D\), GNN consistently outperforms RF. The best gain is at \(100D\), where MSE drops from \(0.25\) to \(0.19\), a relative improvement of \(\mathbf{24\%}\). At \(500D\), the gain is \(\mathbf{15.8\%}\).

\begin{table}[h]
    \small
    \centering
    \caption[The $MSE$ scores of the GNN and RF regression models for predicting the performance of modular algorithm variants for the BBOB problem instances.]{The $MSE$ scores of the GNN and RF regression models for predicting the performance of modular algorithm variants for the BBOB problem instances in 5 and 30 dimensions.}

    \begin{tabular}{
    |p{0.045\textwidth}|
    >{\centering}p{0.03\textwidth}>{\centering}p{0.03\textwidth}|
    >{\centering}p{0.03\textwidth}>{\centering}p{0.03\textwidth}|
    >{\centering}p{0.03\textwidth}>{\centering}p{0.03\textwidth}|
    >{\centering}p{0.03\textwidth}>{\centering\arraybackslash}p{0.03\textwidth}|}
\hline
\multirow{2}{*}{Budget} & \multicolumn{2}{c|}{CMA-ES 5D} & \multicolumn{2}{c|}{CMA-ES 30D} & \multicolumn{2}{c|}{DE 5D} & \multicolumn{2}{c|}{DE 30D} \\
\cline{2-9}
 & GNN & RF & GNN & RF & GNN & RF & GNN & RF \\
\hline
$50D$     & \textbf{0.75} & 0.78 & \textbf{0.15} & \textbf{0.15} & \textbf{0.36} & 0.37 & \textbf{0.21} & 0.26 \\
$100D$    & \textbf{1.16} & 1.22 & \textbf{0.19} & 0.27          & \textbf{0.39} & 0.43 & \textbf{0.19} & 0.25 \\
$300D$    & \textbf{3.68} & 3.98 & \textbf{0.75} & 1.03          & \textbf{0.62} & 0.81 & \textbf{0.27} & 0.30 \\
$500D$    & \textbf{4.39} & 4.85 & \textbf{0.85} & 1.27          & \textbf{1.00} & 1.04 & \textbf{0.32} & 0.38 \\
$1000D$   & \textbf{4.38} & 5.22 & \textbf{1.09} & 1.72          & 2.08          & \textbf{1.95} & \textbf{0.49} & 0.54 \\
$1500D$   & \textbf{4.57} & 5.19 & \textbf{1.34} & 1.88          & \textbf{2.26} & 2.34 & \textbf{0.58} & 0.63 \\
\hline
\end{tabular}
\label{tab:GNN_MSE_scores_regression_models}
\end{table}

\section{Conclusions}
\label{sec:conclusions}
In this study, we explored the use of GNNs for predicting the performance of modular optimization algorithms in numerical black-box optimization. Our results show that GNNs outperform tabular RF models by capturing relationships between problems, algorithm configurations, and performance outcomes. To our knowledge, this is the first work to apply heterogeneous graph structures and GNNs to performance prediction in this domain, offering a promising direction for enhancing meta-learning tasks such as algorithm configuration and selection.

Promising directions for future research include exploring alternative GNN architectures beyond GraphSAGE, such as graph attention networks~\cite{velickovic2017graph} or graph transformers~\cite{yun2019graph}, which could further enhance predictive performance. Explainability methods like GNNExplainer~\cite{ying2019gnnexplainer} can provide insights into which nodes, relations, and features most influence predictions. Leveraging pretrained GNNs through fine-tuning on our heterogeneous graphs may also improve generalization. A key challenge ahead is extending this approach to non-modular algorithms, which requires a standardized vocabulary for algorithmic components and their configurations -- a task that calls for broader community collaboration.

\begin{acks}
\noindent We acknowledge the support of the Slovenian Research and Innovation Agency via program grants No. P2-0098 and P2-0103, and project grants No.J2-4460 and GC-0001. This work is also funded by the European Union under Grant Agreement 101187010 (HE ERA Chair AutoLearn-SI).
\end{acks}

\bibliographystyle{ACM-Reference-Format}

\end{document}